\documentclass[10pt]{article} 
\usepackage[accepted]{tmlr}

\usepackage{microtype}
\usepackage{graphicx}
\usepackage{subcaption}
\usepackage{booktabs} 

\usepackage{hyperref}

\usepackage{amsmath}
\usepackage{amssymb}
\usepackage{mathtools}
\usepackage{amsthm}

\usepackage{xcolor}
\usepackage{multirow}

\usepackage[capitalize,noabbrev]{cleveref}

\usepackage{amsmath,amsfonts,bm}

\newcommand{\pref}{p_{\mathrm{ref}}}

\def\eqref#1{equation~\ref{#1}}

\def\1{\bm{1}}

\DeclareMathAlphabet{\mathsfit}{\encodingdefault}{\sfdefault}{m}{sl}
\SetMathAlphabet{\mathsfit}{bold}{\encodingdefault}{\sfdefault}{bx}{n}

\usepackage{url}

\title{Epipolar Geometry Improves Video Generation Models}

\author{
\name Orest Kupyn \\
\addr University of Oxford
\AND
\name Th\'eo Uscidda \\
\addr Google Research \\
CREST-ENSAE, Institut Polytechnique de Paris
\AND
\name Marta Tintore Gazulla \\
\addr Google Research
\AND
\name Fabian Manhardt \\
\addr Google Research
\AND
\name Federico Tombari \\
\addr Google Research \\
Technical University of Munich
\AND
\name Christian Rupprecht \\
\addr University of Oxford
}

\def\month{11}  
\def\year{2026} 
\def\openreview{\url{https://openreview.net/forum?id=xKvQakeoBP}} 

\begin{document}

\maketitle

{\small\noindent\textbf{Project page:} \url{https://epipolar-dpo.github.io/}\par}

\begin{abstract}
Video generation models have advanced significantly through the latent diffusion transformers trained with rectified flow techniques. Yet these models still struggle with geometric inconsistencies, unstable motion, and visual artifacts that break the illusion of realistic 3D scenes. 3D-consistent video generation could significantly impact numerous downstream applications in generation and reconstruction tasks.
We explore how epipolar geometry constraints improve modern video diffusion models. Despite using massive training data, these models fail to capture fundamental geometric principles. We align diffusion models using pairwise epipolar geometry constraints via preference-based optimization, directly addressing unstable trajectories and geometric artifacts through mathematically principled geometric enforcement.
Our approach efficiently enforces geometric principles without requiring end-to-end differentiability. Evaluation demonstrates that classical geometric constraints provide more reliable preference signals than the video-quality and geometry metrics based on foundational models. We construct preferences from static scenes with camera motion, where the epipolar metric is most reliable, while the aligned model generalizes to dynamic scenes. By bridging data-driven learning with classical computer vision, we reduce epipolar error by 31\% and improve human-rated consistency from 54\% to 72\% without compromising visual quality.
\end{abstract}
\section{Introduction}

{\looseness
=-1
Video generation has witnessed remarkable progress, with recent models~\citep{Sora, veo3_reasoning, moviegen, wan, hunyuanvideo} producing increasingly realistic content from text and image conditions. This advancement has spurred researchers to repurpose these powerful video models for broader applications, including animation~\citep{hi3d}, virtual worlds generation~\citep{he2025cameractrl}, and novel view synthesis~\citep{zhou2025stable}.
}
{\looseness=-1
Video diffusion models are trained on vast volumes of data, developing a strong understanding of object appearance, motion patterns, and scene composition. Many recent works aim to utilize these priors in various downstream tasks~\citep{geo4d, sv3d, v3d}. Despite this, these models still struggle to maintain perfect 3D consistency, often producing content with imperfect geometry, unstable motion, and perspective flaws, even though almost all of the training data is 3D-consistent. Some approaches for enhancing 3D consistency rely on noise optimization~\citep{equivdm}, explicit guidance through point clouds~\citep{zhang2024world, hou2024training}, or camera parameters~\citep{cami2v}. Nevertheless, noisy control signals can constrain the model's generation capabilities, and latent-space optimization makes it difficult to compute direct geometric losses.
}

{\looseness=-1
With the rising popularity of reinforcement learning for model alignment~\citep{dpo, grpo, rlhf}, post-training alignment has gained attention in diffusion model research. Methods such as VideoReward~\citep{videoreward} finetune vision-language models on human preference data, enabling direct supervision through reward models. However, human-annotated quality scores introduce noisy signals and are expensive to collect. Human judgments are inherently subjective and may not capture geometric principles ensuring 3D consistency. The gap between subjective human evaluations and objective geometric requirements creates an opportunity for alignment methods that leverage more mathematically grounded metrics for video quality assessment.
}

\begin{figure*}[t]
\centering
\includegraphics[width=\textwidth]{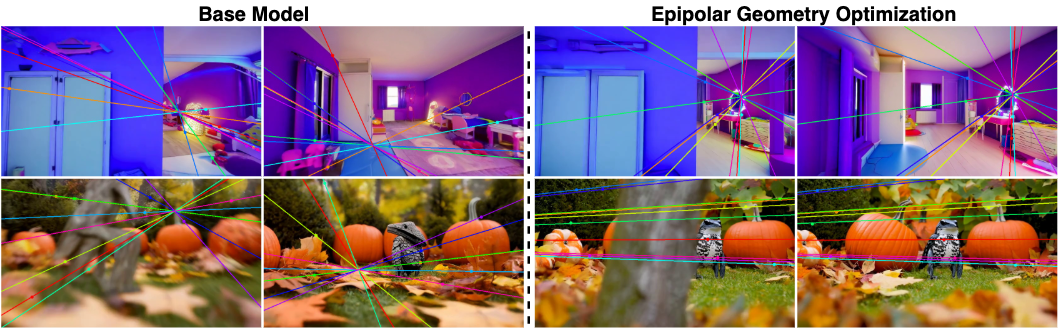}
\caption{\textbf{Improving 3D Consistency:} First and middle frames from the same generated scenes. As the viewpoint changes, the base model exhibits distortions, and visual artifacts, whereas epipolar geometry optimization preserves the structure of the scene and produces more consistent camera motion.}
\label{fig:example}
\end{figure*}

{\looseness=-1
We propose a simple approach that bridges modern video diffusion models with classical computer vision algorithms. Rather than incorporating explicit 3D guidance during generation, we use well-established non-differentiable geometric constraints as reward signals in a preference-based finetuning framework. Specifically, we leverage epipolar geometry constraints to assess 3D consistency between frames. By sampling multiple videos conditioned on the same prompt, we generate diverse camera trajectories that vary in geometric coherence. Epipolar geometry metrics provide reliable signals for identifying which generations better adhere to the principles of projective geometry, enabling us to rank videos and to create training pairs that guide the model toward improved geometric consistency.
}

{\looseness=-1
Our method implements this through Direct Preference Optimization (DPO)~\citep{dpo}, requiring only relative rankings rather than absolute reward values. This bypasses the difficulties of directly using non-differentiable computer vision algorithms in the training loop. By finetuning the model to prioritize generations that satisfy principled geometric constraints, we guide it towards generating inherently more 3D-consistent videos without restricting creative capabilities or requiring explicit 3D supervision. The aligned model shows enhanced 3D consistency, smoother camera trajectories, and fewer artifacts compared.
}

{\looseness=-1
While simple in nature, this paper shows that a basic geometric constraint, described in 1982~\citep{sampson1982fitting}, can recover what video models fail to do, even after large-scale training on billion-scale data: 3D consistency.
}

In summary, the key contributions are as follows:

{\looseness=-1
\noindent \textbf{Epipolar Geometry Optimization:} We introduce a method for finetuning video diffusion models using epipolar geometry constraints as reward signals, particularly leveraging Sampson distance to enhance 3D video consistency without needing differentiability. The models finetuned with reliable signals from classical vision algorithms achieve superior consistency and quality, significantly reducing artifacts and unstable motion in generated content. Our approach demonstrates that aligning models with fundamental geometric principles enables the generalizable shaping of the model's sample distribution, which is independent of the video content.
}

{\looseness=-1
\noindent \textbf{Analysis of Alignment Challenges:} We analyze two important challenges in optimizing 3D scene consistency: motion-consistency tradeoff and generalization to dynamic scenes, and propose simple yet efficient ways to address both. Finetuning with static motion penalty prevents the degenerate solution of improving consistency by reducing motion, while a combination of implicit KL-divergence, compact trainable adapter, and adaptive timestamp-weighting allows the model to generalize well to dynamic content even when trained on static scenes with dynamic cameras.
}

{\looseness=-1
\noindent \textbf{Comprehensive Evaluation Framework:} We develop an extensive evaluation protocol measuring perceptual quality, 3D consistency, motion stability, visual fidelity, and generalization across diverse scenarios. We compare multiple alignment approaches, demonstrating that classical geometric constraints provide more stable optimization signals than modern learned metrics. In addition, we release a dataset of 162,000 generated videos with geometric consistency annotations. The data and code enable further research in geometry-aware video generation.
}
\begin{figure*}[t]
\centering
\includegraphics[width=\textwidth]{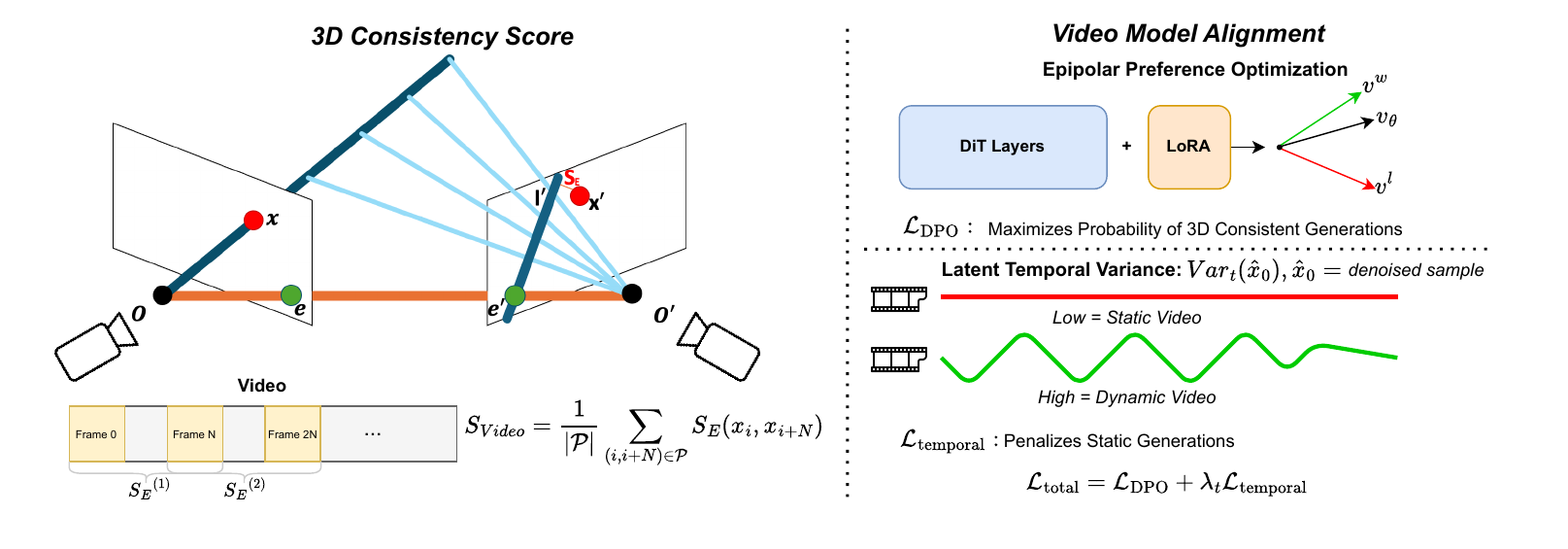}
\caption{\textbf{Epipolar Geometry Optimization.} Left: In 3D-consistent videos, corresponding points must lie on epipolar lines; we compute the distance from matched points to their expected epipolar lines in sampled frame pairs. Right: Flow-DPO training with a static penalty aligns the model toward geometrically consistent outputs.}
\label{fig:pipeline}
\end{figure*}

\section{Background and Related Work}


{\looseness=-1
\textbf{Flow Matching:} Flow Matching~\citep{flow_match} learns a time-dependent vector field that transforms samples from a simple prior (e.g., standard Gaussian) into samples from a target distribution. Given samples from an unknown data distribution $q(x_0)$, the goal is to learn a vector field $v_\theta(x_t, t)$ that generates a probability path $p_t(x)$ interpolating between the prior and data distributions. The Flow Matching objective regresses a neural network against a target vector field $u_t(x)$ that generates this path:
}
\begin{equation}
\mathcal{L}_{\mathrm{FM}}(\theta) = \mathbb{E}_{t, x \sim p_t} \|v_\theta(x, t) - u_t(x)\|^2 \, .
\end{equation}

Modern video diffusion models~\citep{wan} adopt Rectified Flow~\citep{rectified_flow}, defining a linear interpolation between data $x_0 \sim q(x_0)$ and noise $x_1 \sim \mathcal{N}(0, I)$:
\begin{equation}
x_t = (1 - t)x_0 + tx_1, \quad t \in [0, 1] \, .
\end{equation}
This yields a constant target velocity $v = x_1 - x_0$, simplifying the objective to:
\begin{equation}
\mathcal{L}(\theta) = \mathbb{E}_{t, x_0, x_1} \|v - v_\theta(x_t, t)\|^2
\end{equation}
At inference, samples are generated by integrating the learned velocity field from $t=1$ (noise) to $t=0$ (data).

{\looseness=-1
\textbf{Epipolar Geometry:} Epipolar geometry describes the projective relationship between two views of the same scene, depending only on camera parameters and relative pose. When a 3D point $\mathbf{X} \in \mathbb{R}^3$ is observed by two cameras, its projections $\mathbf{x} \in \mathbb{R}^2$ and $\mathbf{x}' \in \mathbb{R}^2$ in the respective image planes are geometrically constrained. The 3D point, camera centers, and both image points must lie on a common plane called the \emph{epipolar plane}. This plane intersects each image along a line called the \emph{epipolar line}: a point at location $\mathbf{x}$ in the first image has its correspondence $\mathbf{x}'$ in the second image lying on the epipolar line $\mathbf{l}' = \mathbf{F}\mathbf{x}$. The fundamental matrix $\mathbf{F} \in \mathbb{R}^{3 \times 3}$ encodes this relationship algebraically using homogeneous coordinates. For corresponding points observing the same 3D point:
}
\begin{equation}
\mathbf{x}'^\top \mathbf{F} \mathbf{x} = 0 \, .
\end{equation}
The Sampson error~\citep{sampson1982fitting} quantifies how well a pair of corresponding points satisfies the epipolar constraint:
\begin{equation}
\label{eq:sampson}
S_E = \frac{(\mathbf{x}'^\top \mathbf{F} \mathbf{x})^2}{\|\nabla_{\mathbf{x}}(\mathbf{x}'^\top \mathbf{F} \mathbf{x})\|^2 + \|\nabla_{\mathbf{x}'}(\mathbf{x}'^\top \mathbf{F} \mathbf{x})\|^2} \, .
\end{equation}
The numerator measures algebraic violation, while the denominator normalizes by the gradient magnitude, providing a first-order approximation to the geometric distance between each point and its epipolar line.

{\looseness=-1
\textbf{Flow Models Alignment:}
Since image and video latent flow models are trained on internet-scale noisy data, efficient finetuning and alignment strategies have emerged. A common approach frames alignment as reinforcement learning, maximizing a reward function while keeping the model close to a reference distribution:
}
\begin{equation}
\label{eq:rlhf_objective}
   \max_{\theta} \mathbb{E}_{x_0 \sim p_\theta}\bigl[r(x_0)\bigr] - \beta \, D_{\mathrm{KL}}\bigl[ p_\theta \Vert \pref \bigr] \, ,
\end{equation}
where $r(x_0)$ is a reward function and $\beta$ controls the KL-divergence regularization strength. Early methods such as DRAFT~\citep{draft} and AlignProp~\citep{alignprop} backpropagate through differentiable reward functions, while DPOK~\citep{dpok} and DDPO~\citep{ddpo} employ policy gradient methods to handle non-differentiable rewards.

{\looseness=-1
Diffusion-DPO~\citep{diffdpo} solves~\cref{eq:rlhf_objective} analytically, eliminating explicit reward model access. Given preference pairs $(x_0^w, x_0^l)$ where $x_0^w$ is preferred over $x_0^l$, define the noise prediction error $\mathcal{E}_\theta(\mathbf{x}_t, t){=}\|\boldsymbol{\epsilon} {-} \boldsymbol{\epsilon}_\theta(\mathbf{x}_t, t)\|^2$. We write $\mathcal{E}_\theta^w$ for the error on the preferred sample and $\mathcal{E}_\theta^l$ for the less preferred; similarly, $\mathcal{E}_{\mathrm{ref}}^{w,l}$ denotes errors under the reference model. The Diffusion-DPO objective is:
}
\begin{equation}\label{eq:diffusion_dpo}
\mathcal{L}_{\mathrm{DPO}} = - \mathbb{E} \Bigl[ \log \sigma \Bigl( -\tfrac{\beta}{2} \bigl( \mathcal{E}_\theta^w - \mathcal{E}_{\mathrm{ref}}^w - \mathcal{E}_\theta^l + \mathcal{E}_{\mathrm{ref}}^l \bigr) \Bigr) \Bigr] ,
\end{equation}
where $\mathbf{x}_t = (1{-}t)\mathbf{x}_0 + t\boldsymbol{\epsilon}$ with $\boldsymbol{\epsilon} \sim \mathcal{N}(0, \mathbf{I})$. This requires only pairwise preferences and enables direct finetuning in latent space.

For rectified flow models~\citep{flow_match, liu2022flow, albergo2022building}, the noise prediction relates to velocity prediction via $\mathcal{E} = (1{-}t)^2 \mathcal{V}$. Define the velocity error $\mathcal{V}_\theta(\mathbf{x}_t, t) {=} \|v {-} v_\theta(\mathbf{x}_t, t)\|^2$ with $v = \boldsymbol{\epsilon} - \mathbf{x}_0$. Using analogous notation $\mathcal{V}_\theta^{w,l}$ and $\mathcal{V}_{\mathrm{ref}}^{w,l}$, and  $\beta_t = \beta(1{-}t^2)$, the Flow-DPO loss~\citep{videoreward} is:
\begin{equation}\label{eq:flow_dpo}
\mathcal{L}_{\mathrm{F\text{-}DPO}} = - \mathbb{E} \Bigl[ \log \sigma \Bigl( -\tfrac{\beta_t}{2} \bigl( \mathcal{V}_\theta^w - \mathcal{V}_{\mathrm{ref}}^w - \mathcal{V}_\theta^l + \mathcal{V}_{\mathrm{ref}}^l \bigr) \Bigr) \Bigr].
\end{equation}

{\looseness=-1
In practice, alignment methods rely on learned reward models to generate preference pairs. Latent image diffusion models~\citep{sdxl, ldm} finetune on data ranked by aesthetics classifiers~\citep{Schuhmann2022LAION}, while VideoReward~\citep{videoreward} trains VLMs on human annotations. However, learned rewards require expensive annotations, capture subjective preferences that may not reflect geometric correctness, and can produce inconsistent signals across different scene types. Recent work explores structured reward signals: DSO~\citep{li2025dso} employs DPO to align 3D generators with physical soundness, and PISA~\citep{pisa} improves the physical stability of video generators using multi-component reward functions. Our method leverages classical computer vision algorithms to provide objective, mathematically grounded preference signals based on epipolar geometry, resulting in more reliable alignment with 3D consistency principles. Since the initial public release of this work, several subsequent papers have expanded the direction of geometry-aware video alignment. VideoGPA \cite{videogpa} apply DPO with a scoring based on foundational geometry models \cite{vggt} ; VIGOR \cite{yin2026vigor} develops a geometry-oriented reward for temporal generative alignment; GeoFlow \cite{ackermann2026geoflow} and VGGRPO \cite{an2026vggrpo} explicitly compute reward for dynamic scenes and extend the alignment to GRPO \cite{grpo}. Geo-Align \cite{geoalign} combines metric-scale reward from MapAnything \cite{keetha2025mapanything} with GRPO to improve camera-controlled video generation. Taken together, they provide additional evidence that enforcing 3D consistency is a useful and increasingly important direction for video generation research. While these works share the goal of enforcing stronger geometric or 3D consistency in generated videos, they differ from our method in the supervision signal, reward construction, and optimization procedure. Our contribution is to demonstrate the effectiveness of simple, reliable epipolar geometric consistency as a basis for preference optimization in video generation.
}

{\looseness=-1
\textbf{Video Generation Models:}
Recent advances in video generation have been dominated by closed-source models trained on proprietary datasets, including Sora~\citep{Sora}, Runway Gen-3~\citep{Runway2024Gen3}, Luma AI~\citep{LumaLabs2024DreamMachine}, Pika~\citep{PikaLabs2024Pika}, and Veo~\citep{DeepMind2024Veo2}. Yet their closed nature limits opportunities for fine-tuning.
}
{\looseness=-1
Open-source alternatives have emerged: Stable Video Diffusion~\citep{stable_video_diffusion}, Hunyuan-Video~\citep{hunyuanvideo}, LTX-Video~\citep{ltx}, and Wan-2.1~\citep{wan}. We use Wan-2.1 (1.3B parameters) for its efficient 3D VAE architecture, which is well-suited for adaptation.
}
{\looseness=-1
These models are trained on enormous data volumes exceeding specific application needs, making domain-aware alignment valuable. V3D~\citep{v3d} finetunes for 3D reconstruction, while VideoReward~\citep{videoreward} introduced RL-based alignment. However, prior methods rely on subjective human preferences or VLMs mimicking them. Our approach optimizes against mathematical rules, aligning models with fundamental geometric principles.
}
\section{Method}
\label{sec:method}
\begin{figure*}[t]
\centering
\includegraphics[width=\textwidth]{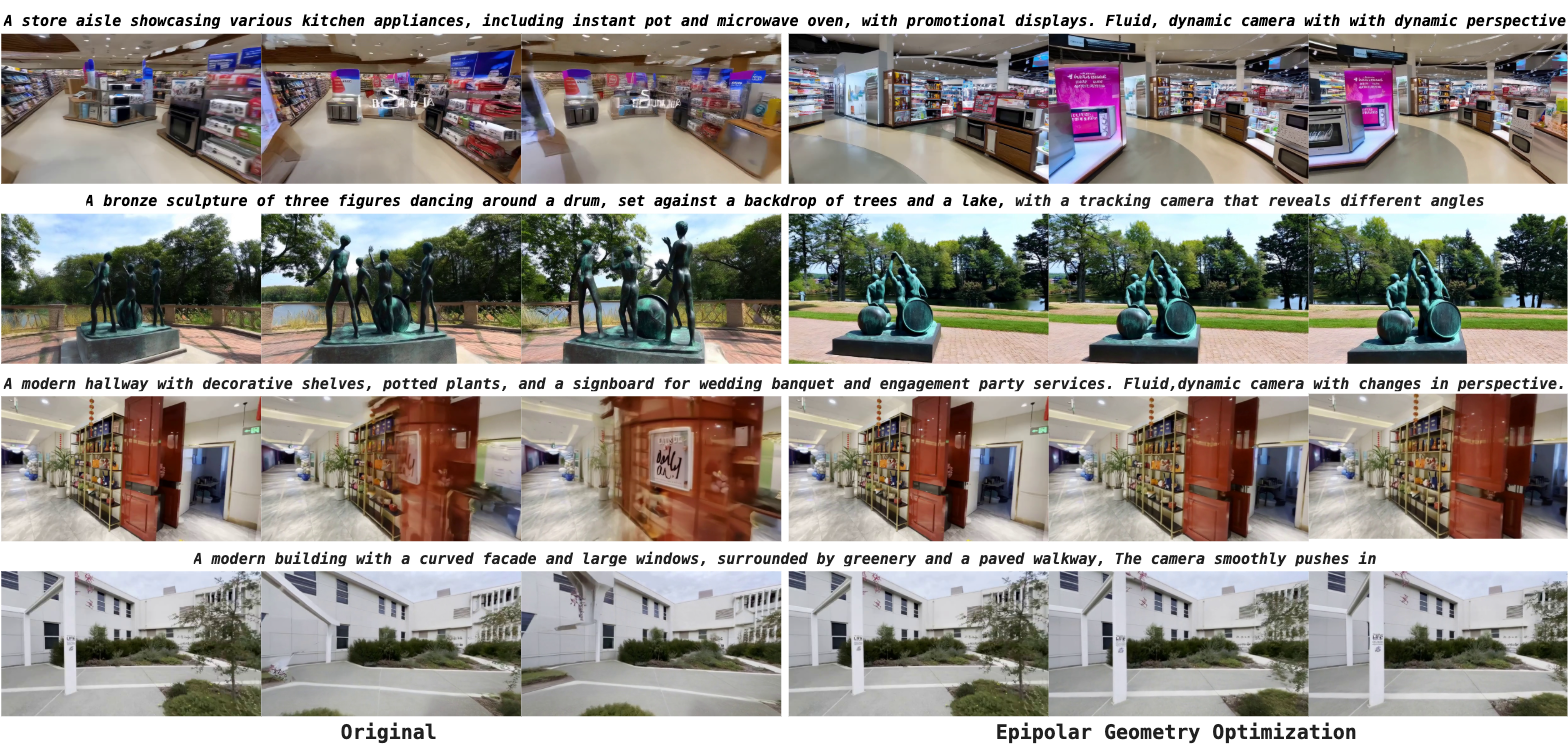}
\caption{\textbf{Qualitative Evaluation:} Each row shows a temporal sequence generated by the base model or the epipolar-aligned model. The first two rows use Wan-2.1-T2V and the last two Wan-2.1-I2V. Across both, the aligned model exhibits fewer distortions and smoother camera motion. Best seen in the videos on the project page \url{https://epipolar-dpo.github.io/}.}
\label{fig:qualitative}
\end{figure*}

{\looseness=-1
We aim to align pretrained video diffusion models to generate geometrically consistent 3D scenes from text or image prompts. We propose an alignment strategy leveraging classical epipolar geometry within a preference-based optimization framework. Traditional reinforcement learning methods~\citep{dppo, grpo} require explicit reward functions and access to final samples, impractical for video models due to non-differentiable rewards and high denoising costs.
Our key observation is that while epipolar geometry constraints do not produce smooth, globally comparable loss surfaces across scene types, the relative intra-prompt error remains consistent. When generating multiple videos with fixed conditioning, diffusion sampling's stochastic nature yields outputs with varying geometric consistency. Epipolar error effectively quantifies relative 3D consistency, with higher values indicating lower consistency.
}
{\looseness=-1
This finding aligns with the direct preference optimization (DPO) paradigm, which requires only relative metrics to determine preference between output pairs rather than absolute reward values. DPO's pairwise comparison nature eliminates the need for globally normalized reward functions, instead leveraging reliable local ranking provided by epipolar geometry measurements to guide model alignment toward more geometrically consistent video generation.
}

\subsection{Objective Function}
Given the pretrained video generator $\pref$ taking text prompt $T$ and optional first frame $I$, generating $x_0 \sim \pref(x_0 | T, I^*)$ where $I^* \in \{I, \emptyset\}$, we want to learn $p_\theta$ optimized for 3D-consistent video sequences. One choice is to optimize:
\begin{equation}
\label{eq:naive_objective}
   \max_{\theta} \mathbb{E}_{c \sim \mathcal{D}_c, x_0 \sim p_\theta}\bigl[r(x_0)\bigr] - \beta \, D_{\mathrm{KL}}\bigl[ p_\theta \Vert \pref \bigr] \, ,
\end{equation}

where $r({x}_0)$ outputs 3D consistency scores, $c{=}(T, I^*)$ is the conditioning, and $\beta$ weights the KL-divergence keeping $p_\theta$ close to $\pref$.
However, $r(x_0)$ relies on non-differentiable algorithms, and evaluating geometric consistency requires complete video generation, which is expensive.

This motivates Flow-DPO (\cref{eq:flow_dpo}), requiring only pairwise preferences. Given dataset $\mathcal{D} {=} \{(c, x_0^w, x_0^l)\}$ where $x_0^w$ has better consistency than $x_0^l$ (lower Sampson error), let $\mathcal{V}_\theta^* {=} \|v^* {-} v_\theta(\mathbf{x}_t^*, t)\|^2$ denote the velocity prediction error. We optimize:
\begin{equation}\label{eq:our_objective}
\mathcal{L} = - \mathbb{E} \Bigl[ \log \sigma \Bigl( -\tfrac{\beta_t}{2} \bigl( \mathcal{V}_\theta^w - \mathcal{V}_{\mathrm{ref}}^w - \mathcal{V}_\theta^l + \mathcal{V}_{\mathrm{ref}}^l \bigr) \Bigr) \Bigr],
\end{equation}
where $\mathbf{x}_t^* = (1{-}t)\mathbf{x}_0^* + t\boldsymbol{\epsilon}^*$, $v^* = \boldsymbol{\epsilon}^* - \mathbf{x}_0^*$, $\beta_t = \beta(1{-}t^2)$.

\textbf{Motion-Consistency Tradeoff:} Static bias is one of the key challenges in optimizing model consistency, significantly increasing the complexity of the task. In theory, a perfectly static video (single image repeated) achieves perfect consistency yet fails as a generative video task.
To prevent degenerate solutions where the model reduces motion to achieve 3D consistency, we add a temporal variation penalty:
\begin{equation}
\mathcal{L}_{\mathrm{temp}} = -\lambda \, \mathbb{E}\bigl[\mathrm{Var}_t(\hat{x}_0)\bigr] \, ,
\end{equation}
where $\hat{x}_0 = x_t + (1{-}t) v_\theta(x_t, t)$ is the predicted clean sample, variance is computed across the temporal dimension, and $\lambda {=} 0.001$. Our final objective is $\mathcal{L}_{\mathrm{total}} = \mathcal{L} + \mathcal{L}_{\mathrm{temp}}$.

Minimizing this loss encourages the model to improve denoising performance on preferred samples $\mathbf{x}_t^w$ relative to less preferred samples $\mathbf{x}_t^l$, guiding the predicted velocity field $v_\theta$ to align with videos exhibiting better 3D consistency while preserving motion quality.

\subsection{Preference Score Computation}

We evaluate the 3D consistency of generated videos by measuring how well they satisfy the epipolar constraints. Given a pair of frames $\mathbf{x}_i$ and $\mathbf{x}_j$ from a generated video, we first compute a set of point correspondences. We use SIFT, a simple and robust handcrafted descriptor \cite{sift}, although the pipeline can also use learned descriptors \citep{lightglue,loftr,xfeat}. These correspondences provide a robust set of matching points between the different viewpoints. We then estimate the fundamental matrix using the normalized 8-point algorithm within a RANSAC \cite{fischler81ransac} framework to handle outliers. The Sampson error (\cref{eq:sampson}) provides a first-order approximation to the geometric distance between a point and its epipolar line. Lower Sampson error values indicate better adherence to projective geometry constraints and, thus, more consistent 3D structure in the generated videos.

To obtain a single consistency score for ranking videos, we uniformly sample every $N$-th frame from a video of length $L$, yielding frames $\{\mathbf{x}_0, \mathbf{x}_N, \mathbf{x}_{2N}, \ldots\}$. We compute the mean Sampson error across consecutive frame pairs:
\begin{align}
\label{eq:video_score}
S_V = \frac{1}{|\mathcal{P}|} \sum_{(i,j) \in \mathcal{P}} S_E(\mathbf{x}_i, \mathbf{x}_j) \, ,
\end{align}
where $\mathcal{P} = \{(0, N), (N, 2N), \ldots\}$ denotes consecutive frame pairs and $S_E(\mathbf{x}_i, \mathbf{x}_j)$ is the mean Sampson error over all point correspondences between frames $i$ and $j$. Lower scores indicate better geometric consistency. For preference pair construction, given videos generated from the same prompt, we assign $x_0^w$ to the video with lower $S_V$ and $x_0^l$ to the one with higher $S_V$.

We analyze the impact of descriptor choice in our ablation studies (\cref{tab:metric_ablation}).

\subsection{Generalization to Dynamic Scenes}
Epipolar constraints assume the relationship between frames can be described by a single fundamental matrix, valid only for static scenes with camera motion. Dynamic objects violate this assumption, corrupting the preference signal. This motivates a key design choice: we train exclusively on static scenes with dynamic cameras, ensuring high geometric consistency metric accuracy.
Despite this, our method generalizes to dynamic content through three mechanisms. First, finetuning only a low-rank adapter (LoRA) \cite{lora} preserves the base model's capacity for diverse dynamic content. Second, implicit KL-divergence regularization in the Flow-DPO objective keeps the finetuned model close to the reference distribution, preventing overfitting to static scene characteristics. Third, we align the model with a timestep-dependent loss weighting that assigns lower weights on early steps $w(t) = \exp\bigl(-2(t - 0.5)^2\bigr)$. This weighting is motivated by the observation that general scene motion is encoded in the first few timestamps \cite{freeinit}, whereas geometric consistency is often violated later, when fine spatial details are generated. Together, these choices allow for the encoding of generalizable modifications to the model's sample distribution that are independent of the video content, as shown in \cref{fig:winrate}.

\subsection{Implementation Details}
We conduct experiments with a state-of-the-art open-source video diffusion model Wan2.1~\citep{wan}, with 1.3 billion parameters.

{\looseness=-1
\textbf{Offline Dataset Generation:} Since epipolar constraints require static scenes with camera motion, we extract text prompts from~\citet{realcam}, which provides diverse indoor and outdoor scene descriptions with dynamic camera movements.
}
{\looseness=-1
To enhance training data quality, we employ Gemma-3 VLM to expand original prompts with more challenging camera motion descriptions, increasing geometric complexity and ensuring sufficient variance in 3D consistency. We generate three videos per caption to increase variation in 3D consistency quality, as preliminary experiments showed pairs from just two samples often lacked meaningful geometric differences. We implement rigorous data filtering: in addition to removing near-static videos, we only sample pairs where $(\mathrm{metric}(x_{\mathrm{win}}) - \mathrm{metric}(x_{\mathrm{lose}}) > \tau) \wedge (\mathrm{metric}(x_{\mathrm{win}}) > \epsilon)$, eliminating pairs where both videos have similar consistency and ensuring we only learn from meaningful gaps. In total, we generate 24,000 triplets for text-to-video and 30,000 triplets for image-to-video training, requiring approximately 1,980 GPU hours on NVIDIA~A6000s.
}

{\looseness=-1
\textbf{Training Configuration:} We implement our approach using Low-Rank Adaptation (LoRA)~\citep{lora} with rank $r{=}64$ and $\alpha{=}128$. This strategy eliminates the need to store the reference model in memory, since the base model with the adapter disabled naturally serves as $\pref$ during training. We train with a batch size of 32 for 10,000 iterations using the AdamW~\citep{adamw} optimizer with a learning rate of $5 \times 10^{-6}$ and 500 warmup steps. The finetuning takes 2~days on 4~A6000 GPUs.
}

\section{Experiments}
\label{sec:experiments}
We evaluate our epipolar-aligned model across three dimensions: 3D consistency, motion stability, and generalization to dynamic scenes, demonstrating that classical geometric constraints provide more reliable optimization signals than learned metrics while improving video generation quality.
\subsection{Evaluation Setup}

{\looseness=-1
We evaluate on 400 videos from DL3DV~\citep{dl3dv} and RealEstate10K~\citep{re10k} test sets, using Gemma-3 VLM~\citep{gemma3} to generate challenging camera motion descriptions. For generalization, we test on VBench~2.0~\citep{vbench}, MiraData~\citep{miradata}, and VideoReward~\citep{videoreward} benchmarks extending beyond static scenes. We measure performance using: (1)~VideoReward VLM for motion quality assessment, (2)~VBench protocol~\citep{vbench} for standardized motion and visual quality metrics, (3)~classical geometric consistency via Sampson epipolar error, and (4)~3D reconstruction quality via Gaussian Splatting to validate downstream task impact.
}

{\looseness=-1
\textbf{Human Evaluation:} We conduct two-stage human evaluation. First, annotators label videos as geometrically consistent based on visible artifacts and motion stability. This reveals that the baseline produces consistent videos only 54.1\% of the time, confirming significant room for improvement. Second, annotators perform pairwise comparisons between baseline and aligned versions, demonstrating that our approach preserves quality for already-consistent content while dramatically improving inconsistent cases (60.4\% vs 7.5\% win rate).
}

\subsection{3D Consistency}


\begin{table*}[htb]
\centering
\caption{\textbf{3D Consistency Evaluation:} Epipolar aligned model improves 3D Scene Reconstruction and is preferred by human evaluators.}
\label{tab:3d_consistency}
\begin{tabular}{l|cc|ccc|c}
\toprule
& \multicolumn{2}{c|}{\textbf{3D Consistency Metrics}} & \multicolumn{3}{c|}{\textbf{3D Scene Reconstruction}} & \textbf{Human Eval} \\
\cmidrule(lr){2-3} \cmidrule(lr){4-6} \cmidrule(lr){7-7}
\textbf{Method} & Sampson Error $\downarrow$ & Perspective Realism $\uparrow$ & PSNR $\uparrow$ & SSIM $\uparrow$ & LPIPS $\downarrow$ & Consistency Rate \\
\midrule
Baseline & 0.190 & 0.426 & 22.32 & 0.706 & 0.343 & 54.1\% \\
Ours & \textbf{0.131} & \textbf{0.428} & \textbf{23.13} & \textbf{0.729} & \textbf{0.315} & \textbf{71.8\%} \\
\bottomrule
\end{tabular}%
\end{table*}
We validate that epipolar geometry alignment improves 3D consistency using three approaches that test different aspects of geometric quality.

{\looseness=-1
\textbf{3D Scene Reconstruction:} We test whether generated videos support accurate 3D scene reconstruction using VGGT~\citep{vggt} to extract scene parameters and camera trajectories. We initialize 3D Gaussian Splatting from extracted scene structure, run 7000 optimization iterations using Splatfacto~\citep{nerfstudio} on 80\% of frames, and evaluate reconstruction fidelity on the remaining 20\%. Our model demonstrates substantial improvements: PSNR increases from 22.32 to 23.13 (+3.6\%), SSIM improves from 0.706 to 0.729 (+3.2\%), and LPIPS decreases from 0.343 to 0.315 ($-$8.2\%). These gains demonstrate that epipolar alignment produces videos with genuinely enhanced 3D structure rather than superficial improvements.
}

{\looseness=-1
\textbf{Geometric Consistency Metrics:} We directly measure adherence to projective geometry principles using classical computer vision algorithms. The Sampson epipolar error shows a dramatic 31\% reduction from 0.190 to 0.131, verifying that our alignment successfully optimizes the metric used for preference selection and confirming that classical epipolar geometry provides clean optimization signals. Additionally, perspective realism, measured by a model trained to evaluate whether image frames contain realistic perspective~\citep{sarkar2024shadows} improves from 0.426 to 0.428, demonstrating positive impact on adjacent geometric metrics despite this metric's inherent noise.
}

\begin{figure}[t]
    \centering

    \begin{subfigure}{0.48\linewidth}
        \centering
        \includegraphics[width=\linewidth]{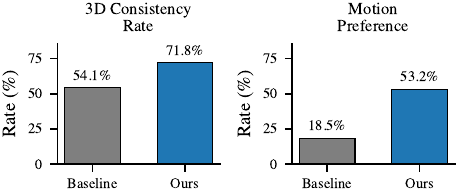}
        \caption{\textbf{Human Evaluation:} Our method improves 3D consistency (54\%$\to$72\% of videos rated consistent) and motion preference (53\% prefer ours vs.\ 19\% baseline in pairwise comparison).}
        \label{fig:human_eval}
    \end{subfigure}
    \hfill
    \begin{subfigure}{0.48\linewidth}
        \centering
        \includegraphics[width=\linewidth]{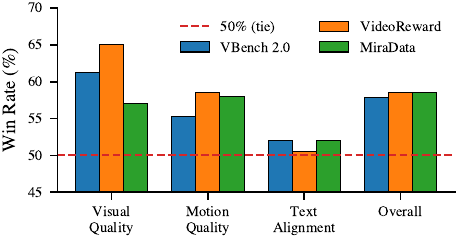}
        \caption{\textbf{Generalization to Dynamic Scenes:} Win rates vs.\ baseline across three benchmarks. Despite training only on static scenes, our method generalizes to dynamic content.}
        \label{fig:winrate}
    \end{subfigure}

    \caption{\textbf{Human Evaluation and Generalization:} The left panel measures human judgments of geometric consistency and motion quality. The right panel evaluates generalization to dynamic scenes.}
    \label{fig:main_results}
\end{figure}

{\looseness=-1
\textbf{Human Evaluation:} While numerical metrics capture geometric aspects, 3D inconsistencies often manifest as artifacts, jitter, or unnatural changes that humans excel at detecting (\cref{fig:human_eval}). Annotators evaluated videos for scene consistency, realism, and artifact-free content. Our method generates significantly more plausible scenes, with 71.8\% of videos labeled as geometrically consistent compared to only 54.1\% for baseline. This 17.7 percentage point improvement demonstrates that geometric alignment benefits are apparent to human observers.
}

\subsection{Motion Quality}
\begin{table*}[htb]
\centering
\caption{\textbf{Motion Quality Evaluation:} Epipolar aligned model improves motion stability and is preferred by human evaluators despite dynamics-consistency tradeoffs.}
\label{tab:motion_quality}
\resizebox{\textwidth}{!}{%
\begin{tabular}{l|ccc|c|c|c}
\toprule
& \multicolumn{3}{c|}{\textbf{VBench Motion Metrics}} & \textbf{VideoReward} & \textbf{Motion Level} & \textbf{Human Eval} \\
\cmidrule(lr){2-4} \cmidrule(lr){5-5} \cmidrule(lr){6-6} \cmidrule(lr){7-7}
\textbf{Method} & Motion Smoothness $\uparrow$ & Dynamic Degree $\downarrow$ & Temporal Flickering $\uparrow$ & Motion Quality $\uparrow$ & Mean SSIM $\downarrow$ & Motion Preference Rate \\
\midrule
Baseline & 0.981 & \textbf{0.751} & 0.958 & 50.0\% & 0.233 & 18.5\% \\
Ours & \textbf{0.984} & 0.710 & \textbf{0.969} & \textbf{69.5\%} & \textbf{0.223} & \textbf{53.2\%} \\
\bottomrule
\end{tabular}%
}
\end{table*}

While geometric alignment enforces smooth, consistent motion and reduced jitter, we also verify that our alignment preserves the model's ability to generate diverse motions. We evaluate motion quality using various metrics, focusing on different aspects, including temporal dynamics, perceptual assessment, and human preference.

{\looseness=-1
\textbf{Temporal Dynamics:} VBench motion metrics show mixed results reflecting the dynamics-consistency tradeoff. Motion smoothness and temporal stability improve, indicating more stable frame-to-frame transitions. However, dynamic degree decreases from 0.751 to 0.710, suggesting reduced motion amplitude. Mean SSIM between first and remaining frames decreases as well, confirming the model's ability to generate dynamic scenes. While we acknowledge the tradeoff, single neural network metrics can exhibit bias, motivating multi-metric evaluation.
}

{\looseness=-1
\textbf{Perceptual Quality Assessment:} VideoReward motion quality evaluation shows substantial improvement with our method, achieving a 69.5\% win rate. This human-distilled assessment validates that geometric consistency training produces motion aligning better with human preferences for natural, stable dynamics, despite some reduction in motion amplitude.
}

{\looseness=-1
\textbf{Human Preference:} Direct human evaluation further validates the motion quality of our method, with annotators preferring our approach at 53.2\% rate across all video types. Since annotators focus on jitter and motion artifacts, our high preference rate shows that stability improvements outweigh motion amplitude reductions. This preference is particularly strong in initially inconsistent videos (60.4\% vs 7.5\% for baseline).
}

\subsection{Supervised Finetuning Evaluation}

We compare the effectiveness of epipolar-aware alignment with standard video model finetuning on multi-view static datasets \cite{dl3dv, re10k}. Directly finetuning the model on perfectly 3D-consistent videos of static scenes could be seen as an alternative approach to improve generation consistency. Yet this naive approach only trains the model to reproduce observed frames, thereby optimizing average risk. It is not directly optimizing task-level objectives, such as multi-view geometric consistency, and instead might replicate biases present in the dataset and regress toward ``mean'' solutions for ambiguous problems. We observe that models finetuned with this objective overfit to dataset-specific factors, such as a narrow set of camera trajectories in RealEstate-10K and DL3DV videos and specific scene types, which hurts the generalization capabilities of the base model. \Cref{tab:multiview_comparison} demonstrates that a model directly finetuned on multi-view datasets does not generalize well to dynamic scenes and, on average, tends to amplify camera motion, mimicking the average training set trajectories.

\begin{table*}[t]
  \caption{\textbf{Comparison vs. Supervised Finetuning on Multi-View Data.} Finetuning on real multi-view static scenes \cite{re10k, dl3dv}  propagates dataset bias such as scene type or specific camera motion. As a result the model overfits to dataset specifics rather than learning general principles.}
  \label{tab:multiview_comparison}
  \centering
  \resizebox{\textwidth}{!}{
  \begin{tabular}{llcccc}
    \toprule
    Benchmark & Method & Visual Quality & Motion Quality & Text Alignment & Overall \\
    \midrule
    \multirow{2}{*}{VideoReward} 
    & Supervised Finetuning & 37.25\% & 46.75\% & \textbf{53.0\%} & 44.75\% \\
    & DPO-Epipolar & \textbf{65.0\%} & \textbf{58.5\%} & 50.5\% & \textbf{58.5\%} \\
    \midrule
    \multirow{2}{*}{VBench} 
    & Supervised Finetuning & 35.8\% & 53.0\% & 39.4\% & 35.2\% \\
    & DPO-Epipolar & \textbf{61.3\%} & \textbf{55.3\%} & \textbf{52.0\%} & \textbf{57.9\%} \\
    \midrule
    \multirow{2}{*}{MiraData9K} 
    & Supervised Finetuning & 41.5\% & 42.0\% & 51.0\% & 47.0\% \\
    & DPO-Epipolar & \textbf{57.0\%} & \textbf{58.0\%} & \textbf{52.0\%} & \textbf{58.5\%} \\
    \bottomrule
  \end{tabular}
  }
\end{table*}

\begin{figure*}[t]
\centering
\includegraphics[width=\textwidth]{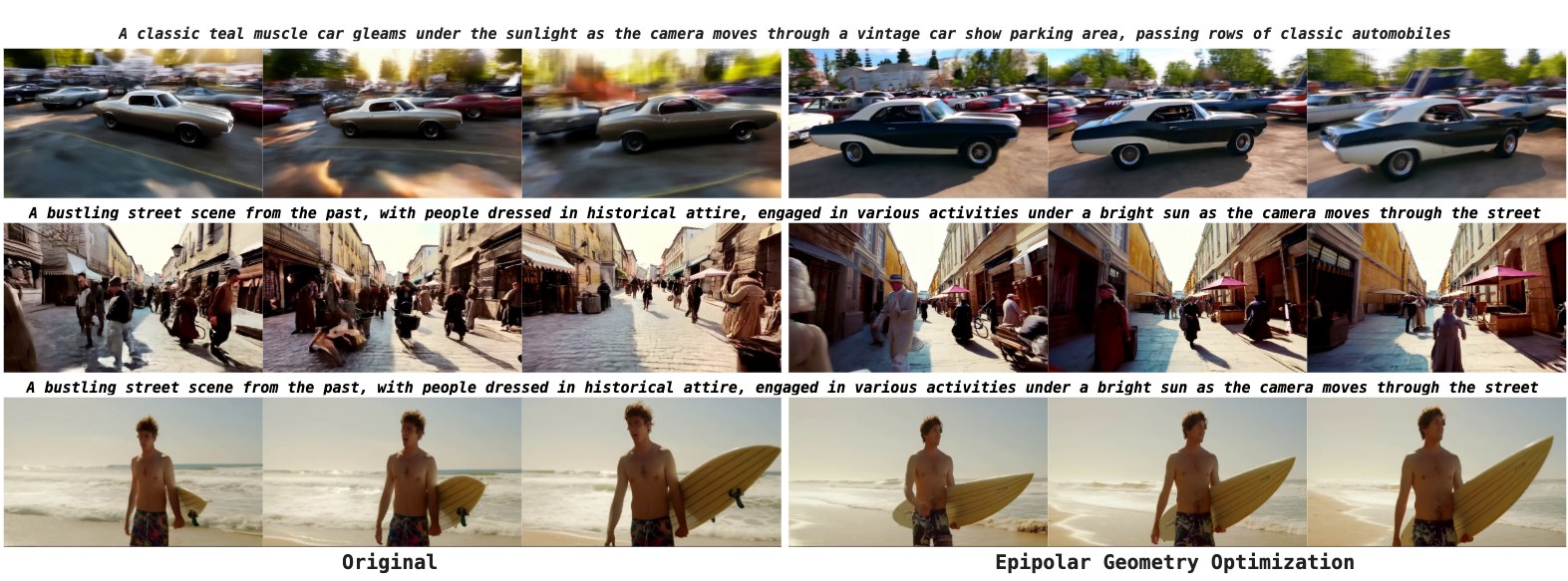}
\caption{\textbf{Generalization to Dynamic Scenes:} The three examples contain both camera motion and independently moving objects: a moving car, a historical street scene, and a person carrying a surfboard. Despite being trained with preferences from static scenes, the epipolar aligned model produces less distortions and more stable trajectories. Best seen in the videos on the project page \url{https://epipolar-dpo.github.io/}.}
\label{fig:qualitative_dynamic}
\end{figure*}

\subsection{Generalization}

Our approach demonstrates strong generalization beyond its training domain of static scenes (\cref{fig:winrate}). Despite training only on static camera motion, our method achieves consistent improvements across VBench~2.0, VideoReward, and MiraData benchmarks, with overall win rates of 57.9\%, 58.5\%, and 58.5\%, respectively.

This generalization occurs because aligning models with geometrically consistent camera trajectories improves video quality even when objects move independently. The primary error sources in dynamic scenes: unstable trajectories, artifacts, and flickering, are amplified by object movement. By learning to produce stable camera motion and reducing geometric inconsistencies, our approach addresses these fundamental issues. VBench metrics confirm that geometric consistency training effectively transfers across diverse scenarios, with improvements in background consistency, temporal stability, and motion smoothness validating our core insight that classical geometric constraints enhance overall 3D understanding.

\subsection{Ablation Study}

\begin{table*}[htb]
  \caption{Win-rate on the VideoReward benchmark comparing different finetuning strategies with geometric consistency metrics and VLM-based metrics (Visual Quality, Motion Quality and Text Alignment).}
  \label{tab:ablation}
  \centering
  \resizebox{\textwidth}{!}{%
  \begin{tabular}{lcccc|ccc}
    \toprule
    & \multicolumn{4}{c|}{\textbf{VideoReward Metrics}} & \multicolumn{3}{c}{\textbf{Consistency Metrics}} \\
    \cmidrule(lr){2-5} \cmidrule(lr){6-8}
    Method & VQ & MQ & TA & Overall & Perspective $\uparrow$ & Sampson $\downarrow$ & Dynamics $\downarrow$ \\
    \midrule
    SFT & \underline{66.0\%} & \underline{63.0\%} & 54.0\% & \underline{64.5\%} & 0.427 & 0.161 & 0.225 \\
    Flow-RWR \cite{videoreward} & 63.5\% & 60.5\% & \textbf{57.0\%} & 64.0\% & \textbf{0.434} & 0.174 & 0.229 \\
    DRO \cite{li2025dso} & 65.0\% & 54.0\% & 50.5\% & \underline{64.5\%} & 0.410 & \textbf{0.068} & \textbf{0.195} \\
    Epipolar-DPO (Ours) & \textbf{72.0\%} & \textbf{71.0\%} & \underline{55.0\%} & \textbf{73.0\%} & \underline{0.428} & \underline{0.127} & \underline{0.223} \\
    \bottomrule
  \end{tabular}%
  }
\end{table*}

We ablate descriptor choices, geometric metrics, alignment methods, and design components. For efficiency the ablations are done on a subset of data.

\begin{figure}[t]
    \centering
    \includegraphics[width=0.68\linewidth]{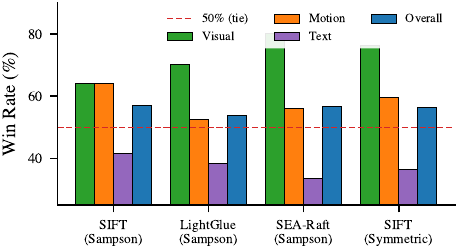}

    \caption{\textbf{Descriptor ablation.}
    Win rate versus the baseline when constructing geometric preferences with different feature descriptors. SIFT with Sampson error provides balanced performance across visual quality, motion quality, text alignment, and overall preference, whereas SEA-RAFT achieves higher visual-quality scores but exhibits reward-hacking behavior.}
    \label{fig:ablation_results}
\end{figure}

{\looseness=-1
\textbf{Descriptor and Metric Analysis:} While SEA-Raft achieves highest visual quality (80.3\%), it hacks the reward by preferring oversaturated scenes. LightGlue finds correspondences in clean areas when videos contain artifacts, resulting in misleadingly low epipolar error, whereas we want correspondences across the entire scene so artifacts anywhere produce high error. Unlike SEA-RAFT, we do not observe comparable reward hacking with SIFT and Sampson error. The metric-noise analysis in \cref{tab:metric_robustness_noise} further shows that pairwise rankings remain robust to moderate pixelwise shift, suggesting that the preference signal is harder to exploit through estimator noise. All setups are comparable, but classical geometric constraints provide cleaner optimization signals than sophisticated alternatives that can miss global inconsistencies.
}

{\looseness=-1
\textbf{Comparison with Learnable Metrics:} 

\begin{table}[htb]
\centering
\caption{\textbf{Metric Comparison:}
SIFT + Sampson outperform VideoReward and MET3R trained models. On the separate VGGT comparison, it achieves comparable perceptual scores and lower Sampson error than VGGT reprojection.}
\label{tab:learned_metrics}
\begin{tabular}{lccc}
\toprule
Preference Metric & VideoReward $\uparrow$ & MET3R $\downarrow$ & Sampson $\downarrow$ \\
\midrule
Motion Quality \cite{videoreward} & 61.3\% & -- & 0.179 \\
MET3R & -- & 0.049 & 0.176 \\
SIFT + Sampson (ours) & 64.3\% & 0.049 & 0.131 \\
\midrule
\multicolumn{4}{l}{\textit{VGGT comparison on the 5K subset}} \\
\midrule
Preference Metric & VQ $\uparrow$ & MQ $\uparrow$ & Sampson $\downarrow$ \\
\midrule
VGGT reprojection & 53.7\% & 56.3\% & 0.105 \\
SIFT + Sampson (ours) & 54.2\% & 55.8\% & 0.098 \\
\bottomrule
\end{tabular}
\end{table}

Table~\ref{tab:learned_metrics} compares learned video-quality and geometry-based preference signals. Models trained with VideoReward Motion Quality achieve a VideoReward score of 61.3\% and a Sampson error of 0.179, whereas our method achieves 64.3\% and 0.131, respectively. Models trained with MET3R achieve a MET3R score of 0.049 but a Sampson error of 0.176; our method matches the MET3R score while achieving substantially lower Sampson error. On the separate VGGT subset, our method performs comparably on perceptual metrics and achieves a slightly lower Sampson error.
}

{\looseness=-1
\textbf{Static Penalty Analysis:} The temporal variation penalty achieves superior motion dynamics (dynamic degree 0.710 vs 0.627) while maintaining comparable geometric consistency and motion quality. This component effectively prevents degenerate static solutions while preserving the geometric alignment. We observe that large $\lambda$ causes the model to always increase camera motion, hurting both 3D consistency and generalization. Similarly, decreasing $\lambda$ forces the model to optimize for a naive solution of increasing 3D consistency by reducing motion.
}

\begin{table}[t]
\centering
\caption{\textbf{Temporal Penalty Weight:} Moderate penalty weights improve dynamic degree while preserving motion quality and geometric consistency. Excessively large weight significantly degrades visual quality.}
\label{tab:temporal_penalty}
\begin{tabular}{c|ccc}
\toprule
\textbf{Penalty Weight $\lambda$} & \textbf{Dynamic Degree} & \textbf{Motion Quality} & \textbf{Sampson Error} \\
\midrule
0 & 0.627 & 71.5\% & 0.127 \\
0.1 & 0.860 & 23.5\% & 0.192 \\
$10^{-5}$ & 0.629 & 70.0\% & 0.130 \\
$10^{-3}$ & 0.710 & 69.5\% & 0.131 \\
\bottomrule
\end{tabular}%
\end{table}

{\looseness=-1
\textbf{Alignment Method Comparison:} Our approach outperforms all alternatives with highest win rates on VideoReward (\cref{tab:ablation}), validating effectiveness of DPO for video model optimization. We compare the alignment against two simple methods: Supervised Finetuning on Win samples, and Flow-RWR~\citep{videoreward} which weights the loss with the normalized reward. All strategies are trained with the same number of steps and target metric. We also find DRO~\citep{li2025dso} interesting as it does not require regularization with the reference model, but in our case, it quickly leads to degenerate solutions deviating from the base model and producing strong artifacts. The other two setups, while still effective, can not learn from the gap between consistent and inconsistent trajectories, resulting in less 3D consistent outputs. The evaluation also show the need to evaluate the model with multiple metrics. For example, learnable metrics are subject to overfitting: overfitted model still can get high scores from a VLM-based metric. 
} 
\section{Limitations}
\label{sec:limitations}
Our approach primarily focuses on static scenes with dynamic camera movements, aligning well with applications in 3D reconstruction and novel view synthesis. Adapting this method to scenes with dynamic objects would require modifying the training pipeline to separately model and evaluate object motion and camera movement. Additionally, epipolar geometry constraints assume point correspondences coming from a static scene under camera motion, limiting effectiveness for scenes with independent object movement or non-rigid deformations where a single fundamental matrix cannot explain all correspondences.
The data mining relies on feature matching and epipolar geometry scoring. However, it can produce false positives (assigning low error to geometrically inconsistent videos) and false negatives (high error to consistent ones) when scenes have repetitive textures, lack distinctive features, or contain significant motion blur. In addition, while the aligned model significantly improves geometric consistency, it does not solve all failure modes. Complex dynamic scenes with extreme camera motion or highly ambiguous content remain challenging for both the baseline and our approach.
\section{Conclusion}
We present an approach for enhancing 3D consistency in video diffusion models by leveraging epipolar geometry constraints as preference signals. Our work shows that classical geometric constraints provide more stable optimization signals than learned metrics, which produce noisy targets. Training on static scenes generalizes effectively to diverse dynamic content, demonstrating the broad applicability of geometric principles. The resulting models generate videos with fewer geometric inconsistencies and more stable camera trajectories while preserving creative flexibility.
This work shows that classical computer vision algorithms effectively complement deep learning, addressing limitations of purely data-driven methods through mathematically grounded geometric principles.

\section*{Acknowledgments}
This work was partially funded through a Google unrestricted gift. We thank Jelena Bratulic for insightful feedback and for helpful discussions.

\bibliography{paper}

@String(ICLR = {Int. Conf. Learn. Represent.})

@String(ICLR  = {ICLR})

@article{dppo,
  title={Training diffusion models with reinforcement learning},
  author={Black, Kevin and Janner, Michael and Du, Yilun and Kostrikov, Ilya and Levine, Sergey},
  journal={arXiv preprint arXiv:2305.13301},
  year={2023}
}

@article{ltx,
  title={Ltx-video: Realtime video latent diffusion},
  author={HaCohen, Yoav and Chiprut, Nisan and Brazowski, Benny and Shalem, Daniel and Moshe, Dudu and Richardson, Eitan and Levin, Eran and Shiran, Guy and Zabari, Nir and Gordon, Ori and others},
  journal={arXiv preprint arXiv:2501.00103},
  year={2024}
}

@misc{DeepMind2024Veo2,
 author = {{Google DeepMind}},
 title = {Veo 2},
 year = {2024},
 month = {12},
 url = {https://deepmind.google/technologies/veo/veo-2/},
 note = {Accessed: 2024}
}

@misc{Runway2024Gen3,
 author = {{Runway}},
 title = {Gen-3},
 year = {2024},
 month = {06},
 url = {https://runwayml.com/},
 note = {Accessed: 2024}
}

@inproceedings{sift,
  title={Object recognition from local scale-invariant features},
  author={Lowe, David G},
  booktitle={Proceedings of the seventh IEEE international conference on computer vision},
  volume={2},
  pages={1150--1157},
  year={1999},
  organization={Ieee}
}

@inproceedings{vbench,
  title={Vbench: Comprehensive benchmark suite for video generative models},
  author={Huang, Ziqi and He, Yinan and Yu, Jiashuo and Zhang, Fan and Si, Chenyang and Jiang, Yuming and Zhang, Yuanhan and Wu, Tianxing and Jin, Qingyang and Chanpaisit, Nattapol and others},
  booktitle={Proceedings of the IEEE/CVF Conference on Computer Vision and Pattern Recognition},
  pages={21807--21818},
  year={2024}
}

@article{rectified_flow,
  title={Flow straight and fast: Learning to generate and transfer data with rectified flow},
  author={Liu, Xingchao and Gong, Chengyue and Liu, Qiang},
  journal={arXiv preprint arXiv:2209.03003},
  year={2022}
}

@article{pisa,
  title={PISA experiments: Exploring physics post-training for video diffusion models by watching stuff drop},
  author={Li, Chenyu and Michel, Oscar and Pan, Xichen and Liu, Sainan and Roberts, Mike and Xie, Saining},
  journal={arXiv preprint arXiv:2503.09595},
  year={2025}
}

@article{li2025dso,
    title   = {DSO: Aligning 3D Generators with Simulation Feedback for Physical Soundness},
    author  = {Li, Ruining and Zheng, Chuanxia and Rupprecht, Christian and Vedaldi, Andrea},
    journal = {arXiv preprint arXiv:2503.22677},
    year    = {2025}
}

@misc{PikaLabs2024Pika,
 author = {{PikaLabs}},
 title = {Pika 1.5},
 year = {2024},
 month = {10},
 url = {https://pika.art/},
 note = {Accessed: 2024}
}

@misc{LumaLabs2024DreamMachine,
 author = {{LumaLabs}},
 title = {Dream Machine},
 year = {2024},
 month = {06},
 url = {https://lumalabs.ai/dream-machine},
 note = {Accessed: 2024}
}

@misc{Sora,
 author = {{OpenAI}},
 title = {Video generation models as world simulators},
 year = {2024},
 url = {https://openai.com/index/video-generation-models-as-world-simulators/},
 note = {Accessed: 2024}
}

@article{rlhf,
  title={Training language models to follow instructions with human feedback},
  author={Ouyang, Long and Wu, Jeffrey and Jiang, Xu and Almeida, Diogo and Wainwright, Carroll and Mishkin, Pamela and Zhang, Chong and Agarwal, Sandhini and Slama, Katarina and Ray, Alex and others},
  journal={Advances in neural information processing systems},
  volume={35},
  pages={27730--27744},
  year={2022}
}

@article{cami2v,
  title={Cami2v: Camera-controlled image-to-video diffusion model},
  author={Zheng, Guangcong and Li, Teng and Jiang, Rui and Lu, Yehao and Wu, Tao and Li, Xi},
  journal={arXiv preprint arXiv:2410.15957},
  year={2024}
}

@article{hou2024training,
  title={Training-free camera control for video generation},
  author={Hou, Chen and Wei, Guoqiang and Zeng, Yan and Chen, Zhibo},
  journal={arXiv preprint arXiv:2406.10126},
  year={2024}
}

@article{zhang2024world,
  title={World-consistent Video Diffusion with Explicit 3D Modeling},
  author={Zhang, Qihang and Zhai, Shuangfei and Bautista, Miguel Angel and Miao, Kevin and Toshev, Alexander and Susskind, Joshua and Gu, Jiatao},
  journal={arXiv preprint arXiv:2412.01821},
  year={2024}
}

@article{equivdm,
  title={EquiVDM: Equivariant Video Diffusion Models with Temporally Consistent Noise},
  author={Liu, Chao and Vahdat, Arash},
  journal={arXiv preprint arXiv:2504.09789},
  year={2025}
}

@article{v3d,
  title={V3d: Video diffusion models are effective 3d generators},
  author={Chen, Zilong and Wang, Yikai and Wang, Feng and Wang, Zhengyi and Liu, Huaping},
  journal={arXiv preprint arXiv:2403.06738},
  year={2024}
}

@inproceedings{sv3d,
  title={Sv3d: Novel multi-view synthesis and 3d generation from a single image using latent video diffusion},
  author={Voleti, Vikram and Yao, Chun-Han and Boss, Mark and Letts, Adam and Pankratz, David and Tochilkin, Dmitry and Laforte, Christian and Rombach, Robin and Jampani, Varun},
  booktitle={European Conference on Computer Vision},
  pages={439--457},
  year={2024},
  organization={Springer}
}

@article{flow_match,
  title={Flow matching for generative modeling},
  author={Lipman, Yaron and Chen, Ricky TQ and Ben-Hamu, Heli and Nickel, Maximilian and Le, Matt},
  journal={arXiv preprint arXiv:2210.02747},
  year={2022}
}

@article{liu2022flow,
  title={Flow straight and fast: Learning to generate and transfer data with rectified flow},
  author={Liu, Xingchao and Gong, Chengyue and Liu, Qiang},
  journal={arXiv preprint arXiv:2209.03003},
  year={2022}
}

@article{albergo2022building,
  title={Building normalizing flows with stochastic interpolants},
  author={Albergo, Michael S and Vanden-Eijnden, Eric},
  journal={arXiv preprint arXiv:2209.15571},
  year={2022}
}

@article{adamw,
  title={Decoupled weight decay regularization},
  author={Loshchilov, Ilya and Hutter, Frank},
  journal={arXiv preprint arXiv:1711.05101},
  year={2017}
}

@inproceedings{hi3d,
  title={Hi3D: Pursuing High-Resolution Image-to-3D Generation with Video Diffusion Models},
  author={Yang, Haibo and Chen, Yang and Pan, Yingwei and Yao, Ting and Chen, Zhineng and Ngo, Chong-Wah and Mei, Tao},
  booktitle={Proceedings of the 32nd ACM International Conference on Multimedia},
  pages={6870--6879},
  year={2024}
}

@article{geo4d,
  title={Geo4D: Leveraging Video Generators for Geometric 4D Scene Reconstruction},
  author={Jiang, Zeren and Zheng, Chuanxia and Laina, Iro and Larlus, Diane and Vedaldi, Andrea},
  journal={arXiv preprint arXiv:2504.07961},
  year={2025}
}

@inproceedings{xfeat,
  title={Xfeat: Accelerated features for lightweight image matching},
  author={Potje, Guilherme and Cadar, Felipe and Araujo, Andr{\'e} and Martins, Renato and Nascimento, Erickson R},
  booktitle={Proceedings of the IEEE/CVF Conference on Computer Vision and Pattern Recognition},
  pages={2682--2691},
  year={2024}
}

@inproceedings{lightglue,
  title={Lightglue: Local feature matching at light speed},
  author={Lindenberger, Philipp and Sarlin, Paul-Edouard and Pollefeys, Marc},
  booktitle={Proceedings of the IEEE/CVF International Conference on Computer Vision},
  pages={17627--17638},
  year={2023}
}

@article{miradata,
  title={Miradata: A large-scale video dataset with long durations and structured captions},
  author={Ju, Xuan and Gao, Yiming and Zhang, Zhaoyang and Yuan, Ziyang and Wang, Xintao and Zeng, Ailing and Xiong, Yu and Xu, Qiang and Shan, Ying},
  journal={Advances in Neural Information Processing Systems},
  volume={37},
  pages={48955--48970},
  year={2024}
}

@inproceedings{loftr,
  title={LoFTR: Detector-free local feature matching with transformers},
  author={Sun, Jiaming and Shen, Zehong and Wang, Yuang and Bao, Hujun and Zhou, Xiaowei},
  booktitle={Proceedings of the IEEE/CVF conference on computer vision and pattern recognition},
  pages={8922--8931},
  year={2021}
}

@article{zhou2025stable,
  title={STABLE VIRTUAL CAMERA: Generative View Synthesis with Diffusion Models},
  author={Zhou, Jensen Jinghao and Gao, Hang and Voleti, Vikram and Vasishta, Aaryaman and Yao, Chun-Han and Boss, Mark and Torr, Philip and Rupprecht, Christian and Jampani, Varun},
  journal={arXiv preprint arXiv:2503.14489},
  year={2025}
}

@misc{veo3_reasoning,
      title={Video models are zero-shot learners and reasoners}, 
      author={Thaddäus Wiedemer and Yuxuan Li and Paul Vicol and Shixiang Shane Gu and Nick Matarese and Kevin Swersky and Been Kim and Priyank Jaini and Robert Geirhos},
      year={2025},
      eprint={2509.20328},
      archivePrefix={arXiv},
      primaryClass={cs.LG},
      url={https://arxiv.org/abs/2509.20328}, 
}

@inproceedings{freeinit,
  title={Freeinit: Bridging initialization gap in video diffusion models},
  author={Wu, Tianxing and Si, Chenyang and Jiang, Yuming and Huang, Ziqi and Liu, Ziwei},
  booktitle={European Conference on Computer Vision},
  pages={378--394},
  year={2024},
  organization={Springer}
}

@article{vpo,
  title={Vpo: Aligning text-to-video generation models with prompt optimization},
  author={Cheng, Jiale and Lyu, Ruiliang and Gu, Xiaotao and Liu, Xiao and Xu, Jiazheng and Lu, Yida and Teng, Jiayan and Yang, Zhuoyi and Dong, Yuxiao and Tang, Jie and others},
  journal={arXiv preprint arXiv:2503.20491},
  year={2025}
}

@inproceedings{sarkar2024shadows,
  title={Shadows Don't Lie and Lines Can't Bend! Generative Models don't know Projective Geometry... for now},
  author={Sarkar, Ayush and Mai, Hanlin and Mahapatra, Amitabh and Lazebnik, Svetlana and Forsyth, David A and Bhattad, Anand},
  booktitle={Proceedings of the IEEE/CVF Conference on Computer Vision and Pattern Recognition},
  pages={28140--28149},
  year={2024}
}

@article{alignprop,
  title={Aligning text-to-image diffusion models with reward backpropagation},
  author={Prabhudesai, Mihir and Goyal, Anirudh and Pathak, Deepak and Fragkiadaki, Katerina},
  year={2023}
}

@article{ddpo,
  title={Training diffusion models with reinforcement learning},
  author={Black, Kevin and Janner, Michael and Du, Yilun and Kostrikov, Ilya and Levine, Sergey},
  journal={arXiv preprint arXiv:2305.13301},
  year={2023}
}

@inproceedings{dpok,
  title={Reinforcement learning for fine-tuning text-to-image diffusion models},
  author={Fan, Ying and Watkins, Olivia and Du, Yuqing and Liu, Hao and Ryu, Moonkyung and Boutilier, Craig and Abbeel, Pieter and Ghavamzadeh, Mohammad and Lee, Kangwook and Lee, Kimin},
  booktitle={Thirty-seventh Conference on Neural Information Processing Systems (NeurIPS) 2023},
  year={2023},
  organization={Neural Information Processing Systems Foundation}
}

@article{draft,
  title={Directly fine-tuning diffusion models on differentiable rewards},
  author={Clark, Kevin and Vicol, Paul and Swersky, Kevin and Fleet, David J},
  journal={arXiv preprint arXiv:2309.17400},
  year={2023}
}

@misc{Schuhmann2022LAION,
 author = {Christoph Schuhmann},
 title = {LAION-Aesthetics},
 year = {2022},
 url = {https://laion.ai/blog/laion-aesthetics/},
 note = {Accessed: 2023-11-10}
}

@inproceedings{ldm,
  title={High-resolution image synthesis with latent diffusion models},
  author={Rombach, Robin and Blattmann, Andreas and Lorenz, Dominik and Esser, Patrick and Ommer, Bj{\"o}rn},
  booktitle={Proceedings of the IEEE/CVF conference on computer vision and pattern recognition},
  pages={10684--10695},
  year={2022}
}

@article{sdxl,
  title={Sdxl: Improving latent diffusion models for high-resolution image synthesis},
  author={Podell, Dustin and English, Zion and Lacey, Kyle and Blattmann, Andreas and Dockhorn, Tim and M{\"u}ller, Jonas and Penna, Joe and Rombach, Robin},
  journal={arXiv preprint arXiv:2307.01952},
  year={2023}
}

@article{he2025cameractrl,
  title={CameraCtrl II: Dynamic Scene Exploration via Camera-controlled Video Diffusion Models},
  author={He, Hao and Yang, Ceyuan and Lin, Shanchuan and Xu, Yinghao and Wei, Meng and Gui, Liangke and Zhao, Qi and Wetzstein, Gordon and Jiang, Lu and Li, Hongsheng},
  journal={arXiv preprint arXiv:2503.10592},
  year={2025}
}

@article{stable_video_diffusion,
  title={Stable video diffusion: Scaling latent video diffusion models to large datasets},
  author={Blattmann, Andreas and Dockhorn, Tim and Kulal, Sumith and Mendelevitch, Daniel and Kilian, Maciej and Lorenz, Dominik and Levi, Yam and English, Zion and Voleti, Vikram and Letts, Adam and others},
  journal={arXiv preprint arXiv:2311.15127},
  year={2023}
}

@article{hunyuanvideo,
  title={Hunyuanvideo: A systematic framework for large video generative models},
  author={Kong, Weijie and Tian, Qi and Zhang, Zijian and Min, Rox and Dai, Zuozhuo and Zhou, Jin and Xiong, Jiangfeng and Li, Xin and Wu, Bo and Zhang, Jianwei and others},
  journal={arXiv preprint arXiv:2412.03603},
  year={2024}
}

@misc{moviegen,
      title={Movie Gen: A Cast of Media Foundation Models}, 
      author={Adam Polyak and others},
      year={2025},
      eprint={2410.13720},
      archivePrefix={arXiv},
      primaryClass={cs.CV},
      url={https://arxiv.org/abs/2410.13720}, 
}

@article{lora,
  title={Lora: Low-rank adaptation of large language models.},
  author={Hu, Edward J and Shen, Yelong and Wallis, Phillip and Allen-Zhu, Zeyuan and Li, Yuanzhi and Wang, Shean and Wang, Lu and Chen, Weizhu and others},
  journal={ICLR},
  volume={1},
  number={2},
  pages={3},
  year={2022}
}

@article{realcam,
  title={Realcam-vid: High-resolution video dataset with dynamic scenes and metric-scale camera movements},
  author={Zheng, Guangcong and Li, Teng and Zhou, Xianpan and Li, Xi},
  journal={arXiv preprint arXiv:2504.08212},
  year={2025}
}

@inproceedings{dl3dv,
  title={Dl3dv-10k: A large-scale scene dataset for deep learning-based 3d vision},
  author={Ling, Lu and Sheng, Yichen and Tu, Zhi and Zhao, Wentian and Xin, Cheng and Wan, Kun and Yu, Lantao and Guo, Qianyu and Yu, Zixun and Lu, Yawen and others},
  booktitle={Proceedings of the IEEE/CVF Conference on Computer Vision and Pattern Recognition},
  pages={22160--22169},
  year={2024}
}

@article{re10k,
  title={Stereo magnification: Learning view synthesis using multiplane images},
  author={Zhou, Tinghui and Tucker, Richard and Flynn, John and Fyffe, Graham and Snavely, Noah},
  journal={arXiv preprint arXiv:1805.09817},
  year={2018}
}

@article{wan,
  title={Wan: Open and advanced large-scale video generative models},
  author={Wang, Ang and Ai, Baole and Wen, Bin and Mao, Chaojie and Xie, Chen-Wei and Chen, Di and Yu, Feiwu and Zhao, Haiming and Yang, Jianxiao and Zeng, Jianyuan and others},
  journal={arXiv preprint arXiv:2503.20314},
  year={2025}
}

@article{gemma3,
  title={Gemma 3 technical report},
  author={Team, Gemma and Kamath, Aishwarya and Ferret, Johan and Pathak, Shreya and Vieillard, Nino and Merhej, Ramona and Perrin, Sarah and Matejovicova, Tatiana and Ram{\'e}, Alexandre and Rivi{\`e}re, Morgane and others},
  journal={arXiv preprint arXiv:2503.19786},
  year={2025}
}

@inproceedings{nerfstudio,
  title={Nerfstudio: A modular framework for neural radiance field development},
  author={Tancik, Matthew and Weber, Ethan and Ng, Evonne and Li, Ruilong and Yi, Brent and Wang, Terrance and Kristoffersen, Alexander and Austin, Jake and Salahi, Kamyar and Ahuja, Abhik and others},
  booktitle={ACM SIGGRAPH 2023 conference proceedings},
  pages={1--12},
  year={2023}
}

@inproceedings{vggt,
  title={Vggt: Visual geometry grounded transformer},
  author={Wang, Jianyuan and Chen, Minghao and Karaev, Nikita and Vedaldi, Andrea and Rupprecht, Christian and Novotny, David},
  booktitle={Proceedings of the Computer Vision and Pattern Recognition Conference},
  pages={5294--5306},
  year={2025}
}

@article{videoreward,
  title={Improving Video Generation with Human Feedback},
  author={Liu, Jie and Liu, Gongye and Liang, Jiajun and Yuan, Ziyang and Liu, Xiaokun and Zheng, Mingwu and Wu, Xiele and Wang, Qiulin and Qin, Wenyu and Xia, Menghan and others},
  journal={arXiv preprint arXiv:2501.13918},
  year={2025}
}

@article{dpo,
  title={Direct preference optimization: Your language model is secretly a reward model},
  author={Rafailov, Rafael and Sharma, Archit and Mitchell, Eric and Manning, Christopher D and Ermon, Stefano and Finn, Chelsea},
  journal={Advances in Neural Information Processing Systems},
  volume={36},
  pages={53728--53741},
  year={2023}
}

@inproceedings{diffdpo,
  title={Diffusion model alignment using direct preference optimization},
  author={Wallace, Bram and Dang, Meihua and Rafailov, Rafael and Zhou, Linqi and Lou, Aaron and Purushwalkam, Senthil and Ermon, Stefano and Xiong, Caiming and Joty, Shafiq and Naik, Nikhil},
  booktitle={Proceedings of the IEEE/CVF Conference on Computer Vision and Pattern Recognition},
  pages={8228--8238},
  year={2024}
}

@article{grpo,
  title={Deepseekmath: Pushing the limits of mathematical reasoning in open language models},
  author={Shao, Zhihong and Wang, Peiyi and Zhu, Qihao and Xu, Runxin and Song, Junxiao and Bi, Xiao and Zhang, Haowei and Zhang, Mingchuan and Li, YK and Wu, Y and others},
  journal={arXiv preprint arXiv:2402.03300},
  year={2024}
}

@article{sampson1982fitting,
  title={Fitting conic sections to “very scattered” data: An iterative refinement of the Bookstein algorithm},
  author={Sampson, Paul D},
  journal={Computer graphics and image processing},
  volume={18},
  number={1},
  pages={97--108},
  year={1982},
  publisher={Elsevier}
}

@article{fischler81ransac,
  title={RC Bolles Random sample consensus: A paradigm for model fitting with applications to image analysis and automated cartography., 1981, 24},
  author={Fischler, MA},
  journal={DOI: https://doi. org/10.1145/358669.358692},
  pages={381--395}
}

@article{videogpa,
  title={VideoGPA: Distilling Geometry Priors for 3D-Consistent Video Generation},
  author={Du, Hongyang and Ye, Junjie and Cong, Xiaoyan and Li, Runhao and Ni, Jingcheng and Agarwal, Aman and Zhou, Zeqi and Li, Zekun and Balestriero, Randall and Wang, Yue},
  journal={arXiv preprint arXiv:2601.23286},
  year={2026}
}

@article{ackermann2026geoflow,
  title={GeoFlow: Enforcing Implicit Geometric Consistency in Video Generation},
  author={Ackermann, Jan and Cai, Shengqu and Deng, Boyang and Kuang, Zhengfei and Peng, Songyou and Wetzstein, Gordon},
  journal={arXiv preprint arXiv:2605.18365},
  year={2026}
}

@article{an2026vggrpo,
  title={Vggrpo: Towards world-consistent video generation with 4d latent reward},
  author={An, Zhaochong and Kupyn, Orest and Uscidda, Th{\'e}o and Colaco, Andrea and Ahuja, Karan and Belongie, Serge and Gonzalez-Franco, Mar and Gazulla, Marta Tintore},
  journal={arXiv preprint arXiv:2603.26599},
  year={2026}
}

@article{yin2026vigor,
  title={VIGOR: VIdeo Geometry-Oriented Reward for Temporal Generative Alignment},
  author={Yin, Tengjiao and Shi, Jinglei and Guo, Heng and Wang, Xi},
  journal={arXiv preprint arXiv:2603.16271},
  year={2026}
}

@article{keetha2025mapanything,
  title={Mapanything: Universal feed-forward metric 3d reconstruction},
  author={Keetha, Nikhil and M{\"u}ller, Norman and Sch{\"o}nberger, Johannes and Porzi, Lorenzo and Zhang, Yuchen and Fischer, Tobias and Knapitsch, Arno and Zauss, Duncan and Weber, Ethan and Antunes, Nelson and others},
  journal={arXiv preprint arXiv:2509.13414},
  year={2025}
}

@misc{geoalign,
      title={Geo-Align: Video Generation Alignment via Metric Geometry Reward}, 
      author={Zizun Li and Haoyu Guo and Runzhe Teng and Chunhua Shen and Tong He},
      year={2026},
      eprint={2605.23903},
      archivePrefix={arXiv},
      primaryClass={cs.CV},
      url={https://arxiv.org/abs/2605.23903}, 
}
\bibliographystyle{tmlr}

\clearpage
\appendix
\section{Visual Quality Evaluation}

Generating more geometrically consistent scenes with fewer artifacts naturally leads to higher overall visual quality of the generated content. To validate this connection, we evaluate visual fidelity across multiple assessment frameworks.

\textbf{Aesthetic Metrics:} VBench \cite{vbench} visual quality assessment shows consistent improvements across multiple dimensions. For example, background consistency increases from 0.930 to 0.942 and aesthetic quality improves from 0.541 to 0.551. These metrics confirm that geometric training enhances visual stability and perceived quality.

\textbf{Perceptual Assessment:} VideoReward \cite{videoreward} visual quality evaluation demonstrates substantial improvement with a 72.0\% win rate, indicating that human-distilled quality assessment strongly favors our geometrically-aligned approach. This suggests that geometric consistency contributes significantly to overall visual appeal.

\textbf{Human Validation:} Human preference evaluation shows a 52.8\% preference rate for our method's visual quality across all video types, further validating that geometric improvements translate to perceptually superior results that human evaluators can identify and prefer.

\begin{table}[htb]
\centering
\caption{Visual Quality and Aesthetic Fidelity Results (Text-to-Video)}
\label{tab:visual_quality}
\resizebox{\columnwidth}{!}{%
\begin{tabular}{l|cc|c|c}
\toprule
& \multicolumn{2}{c|}{\textbf{VBench Visual Metrics}} & \textbf{VideoReward} & \textbf{Human Eval} \\
\cmidrule(lr){2-3} \cmidrule(lr){4-4} \cmidrule(lr){5-5}
\textbf{Method} & Background Consistency $\uparrow$ & Aesthetic Quality $\uparrow$ & Visual Quality $\uparrow$ & Visual Preference Rate \\
\midrule
Baseline & 0.930 & 0.541 & - & 15.0\% \\
Ours & \textbf{0.942} & \textbf{0.551} & \textbf{72.0\%} & \textbf{52.8\%} \\
\bottomrule
\end{tabular}%
}
\end{table}
\begin{table}[htb]
  \caption{\textbf{Win-rate vs.~Wan-2.1-14B \cite{wan}} on the VideoReward \cite{videoreward} benchmark. The Baseline and Epipolar-Aligned Model contain only 1.3B parameters.}
  \label{tab:videoreward_14b}
  \centering
  \begin{tabular}{lcccc}
    \toprule
    \multicolumn{5}{c}{\textbf{Text-to-Video}} \\
    \midrule
    Method & Visual Quality & Motion Quality & Text Alignment & Overall \\
    \midrule
    Baseline & 13.3\% & 14.4\% & 24.2\% & 8.6\% \\
    DPO-Epipolar & \textbf{18.1\%} & \textbf{21.8\%} & \textbf{25.0\%} & \textbf{13.8\%} \\
    \bottomrule
  \end{tabular}
\end{table}

\section{Image-to-Video Evaluation}

\begin{table*}[t]
\centering
\caption{\textbf{Image-to-Video Alignment:} Despite image conditioning constraints, epipolar alignment shows consistent improvements across multiple metrics.}
\label{tab:image_to_video}
\resizebox{\textwidth}{!}{%
\begin{tabular}{l|cc|ccc|cc|ccccc}
\toprule
& \multicolumn{2}{c|}{\textbf{VideoReward}} & \multicolumn{3}{c|}{\textbf{3D Reconstruction}} & \multicolumn{2}{c|}{\textbf{3D Consistency}} & \multicolumn{5}{c}{\textbf{VBench Metrics}} \\
\cmidrule(lr){2-3} \cmidrule(lr){4-6} \cmidrule(lr){7-8} \cmidrule(lr){9-13}
\textbf{Method} & Visual & Motion & PSNR & SSIM & LPIPS & Motion & Sampson & Background & Aesthetic & Temporal & Motion & Dynamic \\
& Quality & Quality & $\uparrow$ & $\uparrow$ & $\downarrow$ & (SSIM $\downarrow$) & Error $\downarrow$ & Consistency & Quality & Flickering & Smoothness & Degree \\
\midrule
Baseline & - & - & \textbf{21.08} & 0.686 & 0.408 & 0.239 & 0.215 & 0.955 & 0.498 & \textbf{0.981} & 0.992 & \textbf{0.378} \\
Ours & \textbf{51.35\%} & \textbf{56.08\%} & 20.99 & \textbf{0.700} & \textbf{0.377} & \textbf{0.239} & \textbf{0.197} & \textbf{0.955} & \textbf{0.499} & 0.980 & \textbf{0.992} & 0.343 \\
\bottomrule
\end{tabular}%
}
\end{table*}

Image-to-video alignment presents unique challenges due to the strong conditioning signal from the input image. The image conditioning is integrated into intermediate layers of the diffusion process, creating additional constraints that naturally reduce output variance and make alignment more challenging. Despite these limitations, our epipolar geometry optimization demonstrates consistent positive impact across multiple evaluation dimensions.

The 3D reconstruction results validate the geometric improvements: SSIM improves from 0.686 to 0.700, and LPIPS decreases from 0.408 to 0.377. These gains, while more modest than text-to-video results, confirm that enhanced geometric consistency translates to better downstream 3D understanding even under image conditioning constraints. The Sampson epipolar error improvement from 0.215 to 0.197 further validates the effectiveness of classical geometric alignment.

VideoReward metrics show meaningful improvements in motion quality (56.08\% vs 43.92\%) and visual quality (51.35\% vs 48.65\%). VBench metrics remain stable with slight improvements in aesthetic quality, demonstrating that geometric optimization preserves overall generation quality while enhancing 3D consistency.

While the input image provides strong structural guidance, it also constrains the model's ability to adapt toward geometrically optimal solutions. Nevertheless, consistent positive trends across reconstruction, consistency, and quality metrics validate that classical geometric constraints provide reliable optimization signals even in constrained generation scenarios.

\section{Metric Robustness}
\label{app:metric_robustness}

We analyze the robustness of the Sampson distance to feature correspondences and fundamental-matrix estimation errors. Since DPO uses pairwise preferences, we test if perturbations change the relative ranking of videos generated from the same prompt.
We first perturb SIFT correspondences with random noise and measure whether the original pairwise ranking is preserved. Rankings remain stable under small and moderate perturbations, and degrade only under severe corruption. We additionally remove a random 10\% of the correspondences before estimating the fundamental matrix with RANSAC. Across 50 pairs, this preserves 99.3\% of ranking decisions, with only 2.0\% of pairs experiencing a winner flip. These results show that estimation noise has a limited impact on the preference labels after filtering. Preference-pair filtering further reduces the effect of noise by selection only pairs with a meaningful score gap.

\begin{table}[h]
\centering
\caption{\textbf{Robustness to geometric-estimation noise.} Pairwise rankings remain stable under realistic correspondence perturbations. The bootstrap experiment removes 10\% of correspondences before fundamental-matrix estimation.}
\label{tab:metric_robustness_noise}
\begin{tabular}{lc}
\toprule
\textbf{Perturbation} & \textbf{Ranking Agreement} \\
\midrule
Tiny, 1 px (0.2\%) & 98.0\% \\
Small, 20 px (4.8\%) & 98.0\% \\
Medium, 40 px (9.6\%) & 82.0\% \\
Severe, 60 px (14.4\%) & 60.0\% \\
Bootstrap 90\% correspondences & 99.3\% \\
\bottomrule
\end{tabular}%
\end{table}

Next, we compare the score distributions of real static-scene videos, generated winners, and generated losers. Lower Sampson error indicates better geometric consistency. The median scores follow the expected ordering, while the inlier ratios provide a consistent diagnostic of correspondence quality.

\begin{table}[t]
\centering
\caption{\textbf{Score separation across real and generated videos.} Real videos receive the lowest median Sampson error, followed by generated winners and generated losers.}
\label{tab:metric_robustness_groups}
\begin{tabular}{lcc}
\toprule
\textbf{Video Group} & \textbf{Median Sampson Error} & \textbf{Median Inlier Ratio} \\
\midrule
Real static-scene videos & 5.187 & 78.9\% \\
Generated $x_w$ & 7.393 & 65.6\% \\
Generated $x_l$ & 18.937 & 31.3\% \\
\bottomrule
\end{tabular}%
\end{table}

These results show that the gap between real videos and generated videos with high and low consistency is significant. The ranking remains stable under realistic correspondence perturbations and resampling, indicating that the metric also provides reliable preference labels.

\section{Inference-Time Optimization Evaluation}

We further compare our method with VPO \cite{vpo}, a video prompt optimization technique, which is complementary to our approach since it optimizes prompts rather than model weights, as well as simple reward guidance \cite{videoreward}, which does not require finetuning a model.

\subsection{Prompt Optimization Evaluation}

We evaluate VPO alone and in combination with our method. Results are reported in \Cref{tab:vpo}.

\begin{table}[h]
\centering
\footnotesize
\caption{\textbf{Prompt Optimization:} VPO optimizes prompts while our method improves geometry alignment. The two approaches are complementary and can be combined to achieve both high visual quality and geometric consistency.}
\label{tab:vpo}
\resizebox{\columnwidth}{!}{
\begin{tabular}{lccccc}
\toprule
\textbf{Method} & \textbf{Visual Quality} $\uparrow$ & \textbf{Motion Quality} $\uparrow$ & \textbf{Overall} $\uparrow$ & \textbf{Dynamic Degree} $\uparrow$ & \textbf{Motion (mean SSIM)} $\downarrow$ \\
\midrule
Ours & 63.1\% & 65.8\% & 59.1\% & \textbf{0.80} & \textbf{0.211} \\
VPO & 59.1\% & 70.6\% & 82.7\% & 0.65 & 0.235 \\
VPO + Ours & \textbf{67.0\%} & \textbf{71.9\%} & \textbf{83.6\%} & 0.61 & 0.234 \\
\bottomrule
\end{tabular}
}
\end{table}

We observe that VPO tends to reduce camera motion and restructure prompts while optimizing for general video quality. However, such prompt optimization methods can be efficiently combined with geometry-aligned models like ours to simultaneously achieve high visual quality and geometric consistency.

\subsection{Inference-Time Reward Guidance}

We evaluate a simple inference-time reward-guidance baseline based on depth reprojection error. This method does not require finetuning a model, yet it increases inference time by computing and backpropagating through the reward. We apply reward guidance every 5 denoising steps over a total of 20 steps. Results are reported in \Cref{tab:reward_guidance}. Reward guidance provides a simple training-free alternative, but it does trade off generation speed for limited improvements in quality.

\begin{table}[h]
\centering
\footnotesize
\caption{\textbf{Inference-Time Reward Guidance:} Runtime is measured over 20 denoising steps, with reward guidance applied every 5 steps.}
\label{tab:reward_guidance}
\begin{tabular}{lcccc}
\toprule
\textbf{Method} & \textbf{Visual Quality} $\uparrow$ & \textbf{Motion Quality} $\uparrow$ & \textbf{Sampson Error} $\downarrow$ & \textbf{Time (20 Steps)} $\downarrow$ \\
\midrule
Reward Guidance & 52.63\% & 52.37\% & 0.168 & 65.10 \\
Ours & \textbf{72.0\%} & \textbf{69.6\%} & \textbf{0.131} & \textbf{54.2} \\
\bottomrule
\end{tabular}
\end{table}

\section{Scaling Analysis}
To understand how our geometric alignment performs across different model scales, we compare both the baseline and epipolar-aligned 1.3B parameter models against the much larger Wan-2.1-14B model \cite{wan}. As shown in \Cref{tab:videoreward_14b}, while the performance gap remains substantial due to the 14B model's higher resolution (720p) and superior base capabilities, our epipolar alignment helps close this gap meaningfully. The aligned 1.3B model achieves win rates of 18.1\%, 21.8\%, and 25.0\% for Visual Quality, Motion Quality, and Text Alignment respectively, compared to 13.3\%, 14.4\%, and 24.2\% for the baseline 1.3B model.
Notably, the 14B model requires approximately 10× longer inference time than the 1.3B variant, making our alignment approach particularly valuable for applications where computational efficiency is critical. This suggests that geometric consistency improvements can partially compensate for scale limitations, offering a practical path toward better video quality without the computational overhead of significantly larger models.

\section{Qualitative Evaluation}
For a comprehensive qualitative assessment of video quality and geometric consistency, we provide full video sequences and interactive baseline-vs.-ours comparisons on the project page \url{https://epipolar-dpo.github.io/}.

\section*{Impact Statement}

Video generation models may be misused to produce realistic but deceptive content, contributing to the spread of misinformation, political manipulation, and erosion of public trust. Furthermore, the computational resources required to train such models raise environmental concerns and may exacerbate inequalities in access to advanced AI technologies.
Geometry-aware video generation can facilitate various 3D vision tasks, including scene reconstruction, SLAM, and visual odometry. By improving geometric consistency in generated videos, our method produces more realistic and usable synthetic data for training computer vision systems. This advances applications in robotics and autonomous navigation, where accurate spatial understanding is crucial. The integration of classical geometry principles with modern generative models represents a promising direction for enhancing AI systems with stronger physical world understanding.


\end{document}